\documentclass[conference]{IEEEtran}
\usepackage{times}
\usepackage[square,numbers]{natbib}
\usepackage{multicol}
\usepackage[bookmarks=true]{hyperref}
\usepackage{amsfonts}
\usepackage{mathtools}
\usepackage{amsmath}
\usepackage{amssymb}
\usepackage{framed}
\usepackage{balance}
\usepackage{subfigure}
\usepackage{graphics} % for pdf, bitmapped graphics files
\usepackage{lipsum}% http://ctan.org/pkg/lipsum
\usepackage{graphicx}% http://ctan.org/pkg/graphicx
\usepackage{mathptmx} % assumes new font selection scheme installed
\usepackage{algorithm}
\usepackage{amssymb}  % assumes amsmath package installed
\usepackage{algpseudocode}
\usepackage{booktabs}
\usepackage{tabularx}
\usepackage{booktabs}% http://ctan.org/pkg/booktabs
\usepackage{array}% http://ctan.org/pkg/array
\usepackage{adjustbox,lipsum}
\usepackage[table]{xcolor}
\usepackage{nicefrac}
\usepackage{mdwtab}
\usepackage{multirow}
\usepackage{colortbl}
\usepackage{hyperref}
\usepackage{color,soul}
\usepackage{epsfig} % for postscript graphics files
\usepackage{graphics} % for pdf, bitmapped graphics files
\usepackage{float}
\DeclareMathOperator*{\argmax}{arg\,max}  % in your preamble
\begin{document}
% paper title
\title{Information-Theoretic Based Target Search with Multiple Agents}

%temporary version
%\author{Minkyu Kim, Ryan Gupta, and Luis Sentis}

% %TODO: Need to be fixedtemporary version
 \author{
  \IEEEauthorblockN{Minkyu Kim}
  \IEEEauthorblockA{Department of Mechanical Engineering\\
  The University of Texas at Austin\\
  Email: steveminq@utexas.edu}
  \and
  \IEEEauthorblockN{Ryan Gupta}
  \IEEEauthorblockA{Department of Aerospace Engineering\\
  Email: ryan.gupta@utexas.edu}
  \and
  \IEEEauthorblockN{Luis Sentis}
  \IEEEauthorblockA{Department of Aerospace Engineering\\
  Email: lsentis@austin.utexas.edu}
 }

% \author{Minkyu Kim$^{1,3}$, Ryan Gupta$^{1,2}$, and Luis Sentis$^{1,2}$% <-this % stops a space
% \thanks{The authors are with the $^{1}$Human Centered Robotics Lab, the $^{2}$ Department of Aerospace Engineering, and the $^{3}$Department of Mechanical Engineering at the University of Texas at Austin}%
% }
\maketitle
% avoiding spaces at the end of the author lines is not a problem with
% conference papers because we don't use \thanks or \IEEEmembership
\begin{abstract}
This paper proposes an online path planning and motion generation algorithm for heterogeneous robot teams performing target search in a real-world environment. Path selection for each robot is optimized using an information-theoretic formulation and is computed sequentially for each agent. First, we generate candidate trajectories sampled from both global waypoints derived from vertical cell decomposition and local frontier points. From this set, we choose the path with maximum information gain. We demonstrate that the hierarchical sequential decision-making structure provided by the algorithm is scalable to multiple agents in a simulation setup. We also validate our framework in a real-world apartment setting using a two robot team comprised of the Unitree A1 quadruped and the Toyota HSR mobile manipulator searching for a person. The agents leverage an efficient leader-follower communication structure where only critical information is shared.
\end{abstract}

\IEEEpeerreviewmaketitle

\section{Introduction}
  
There has been tremendous attention for search behaviors using single and multiple agent systems. These studies can be classified into various categories, such as offline\cite{milp2012, milp2013, agmon2006constructing, indoorsurvey2018} and online \cite{songcpp2018, song2019, proximitycpp2017, karydis2020, gonzalez2005bsa} coverage path planning, exploration and mapping \cite{caglioti2010,dai2020, kumarplan2015, kumarmap2015, todorov2013, slamsurvey2017, lee2020efficient, filatov2020simplistic}, and search itself \cite{ye1999sensor, shubina2010visual, gobelbecker2011planning, rasouli2016sensor, aydemir2013active, aydemir2011search, zhang2012active}. With improvements to robot mobility, sensing and computing power there have been many successful applications \cite{karydis2020,schmuck2017multi,wang2020multi,scherer2015autonomous,woosley2020multi,bellicoso2018advances,dang2020graph} in real-world settings like moon exploration, search and rescue, and unmanned surveillance with teams of robots made up of UAVs, legged systems and mobile robots. Several of these groups have ongoing work focusing on using multiple agents to improve the performance of their system.
  
This study attempts to solve the problem of target search with a single robot or multiple robots within a given search boundary in real-time fashion. The contribution of this paper is an online search algorithm which can be scaled to n heterogeneous agents for performing a probabilistically optimal search in real-world environments. Because the target search problem can be formulated in varying ways, we first discuss a set of assumptions for which we are solving. Such conditions are prior knowledge of the search region, the presence of a target prediction model, and characteristics of the target (i.e., the target being static or dynamic). Our online planning algorithm accounts for changes to the environment and uses sensor observations to perform a probabilistically optimal continuous search of the region.
  
This work proposes the use of multiple agents to perform the search task defined above for settings in which robot speed and sensor coverage are heterogeneous. The most significant challenge for multi-agent systems in real-world cooperation tasks such as search is robust, quick and efficient communication. A key characteristic of a successful design is minimizing information shared between the agents to maintain efficiency and reduce computational time. This is imperative to ensure that agents receive action commands in real-time.
 
\begin{figure}[t]
\centerline{\includegraphics[width=1.0\columnwidth]{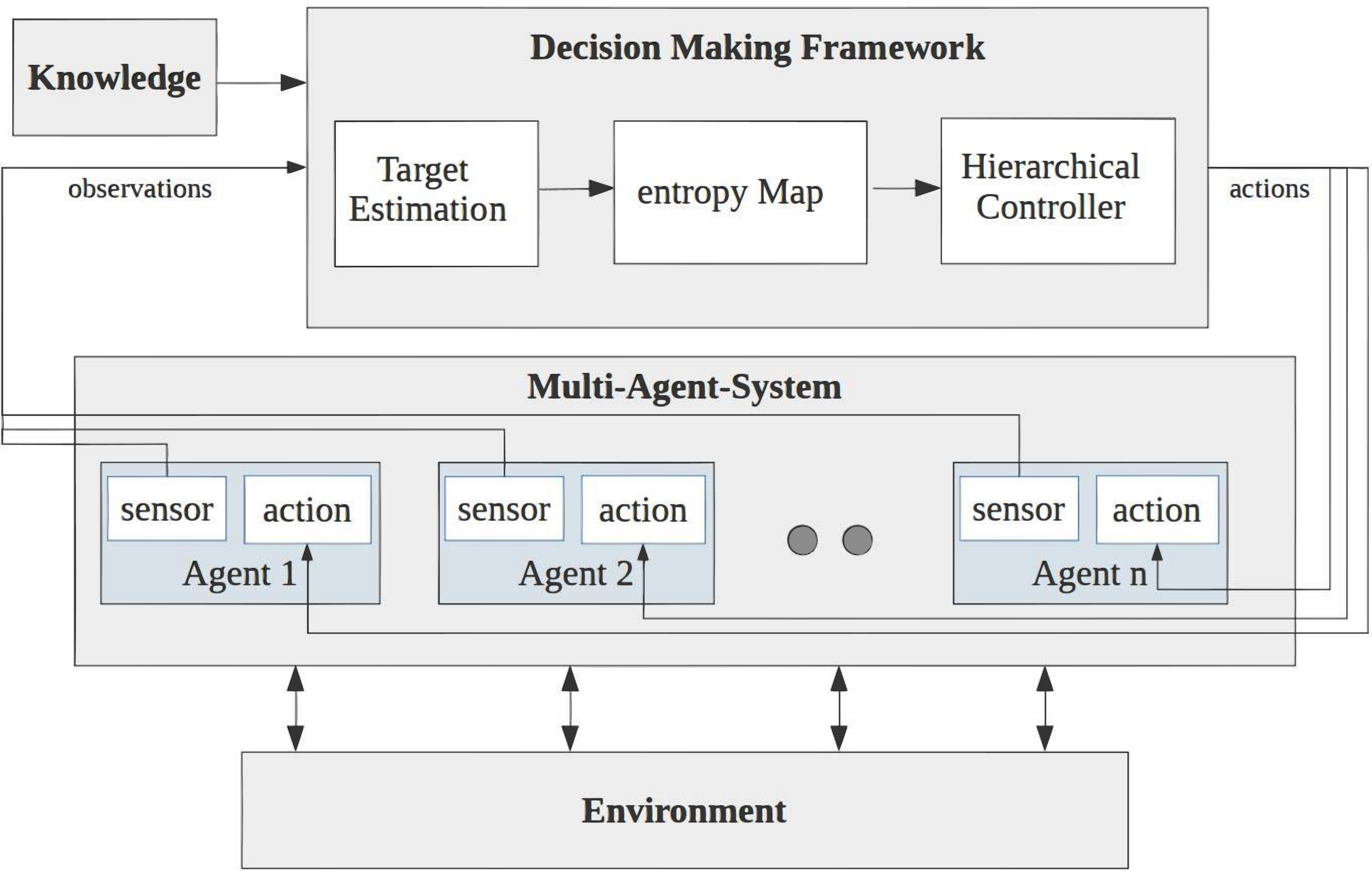}}
\caption{The proposed architecture for Multi-Agent-System }
\label{fig:system}
\end{figure}

\section{Methods}
\subsection{Problem Definition}
We define a cooperative multi-agent target search problem with a hierarchical knowledge structure. This problem aims to find a control policy (sensing action) for multi-agent system (MAS) search in a known area for a target with known probability distribution in the predefined search region. Each agent's control policy is supposed to find control inputs that maximize target information (or reduce target uncertainty). Assuming a heterogeneous robots setup, we denote the index of the agent in each equation.
\subsection{Target Estimation}
We use Bayesian Inference to recursively estimate target state, $x$, through sequential observations, $y$s. Bayesian inference is a commonly used framework used to estimate a target state in a probabilistic manner. This inference model aims to predict the posterior distribution of target position at time $k$, namely, $p(x_k)$. Bayesian filtering includes a prediction step and a correction step using incoming sensing information. Assuming that the prior distribution $p(x_{k-1})$ is available at time $k-1$, the prediction step attempts to estimate $P(x_k|y^{1:n}_{1:k-1})$ \--- where $n$ is the number of agents \--- from previous observations as follows.
\begin{equation}
    p(x_k|y^{1:n}_{1:k-1})=\int p(x_k|x_{k-1})p(x_{k-1}|y^{1:n}_{1:k-1})dx_{k-1},
\label{Eq:1}
\end{equation}
where $p(x_k|x_{k-1})$ is the target's motion model based on a first order Markov process. Then, when the measurement $y^{1:n}_k$ is available, the estimated state can be updated as 
\begin{equation}
    p(x_k|y^{1:n}_k)=\frac{p(y^{1:n}_k|x_k)p(x_k|y^{1:n}_{1:k-1})}{p(y^{1:n}_k|y^{1:n}_{1:k-1})}
\label{Eq:2}
\end{equation}
where $p(y^{1:n}_k|y^{1:n}_{1:k-1})=\int p(y^{1:n}_k|x_k)p(x_k|y^{1:n}_{k-1})dx_k$ and $p(y^{i:n}_k|x_k)$ is a sensing model for multi agent system, which can also be decomposed to each agent's sensing model $p(y^i|x)$. For the correction stage, the measurement of all agents are used to modify the prior estimate, leading to the target belief. If a static target is assumed the target motion model can be described as $p(x_k|x_{k-1})=\mathcal{N}(x_{k-1}; x_k, \Sigma)$, only containing a noise term with the previous target state. Instead if a dynamic target is assumed to have some constant velocity, we can represent the target model as $p(x_k|x_{k-1})= \mathcal{N}(x_{k-1}; x_k+V\Delta , \Sigma)$.

\subsection{Hierarchical Bayesian Model}
We propose a leader-follow hierarchy setting in which we assume that the first agent is the leader and the second is the sub-leader, consisting of sequentially lower-class followers. Assuming that decisions (desired paths) can be mapped into the expected sensing outputs, within the proposed hierarchy structure, our decision-making process can be generalized to $n$-multi agent systems, estimating target state $x_t$ with expected observations of all agents, $[ y^1_{t:t+h}, y^2_{t:t+h}, \cdots, y^n_{t:t+h}]$ with time horizon $h$. To be more specific, given the target state probability $p(x)$ and the expected sensing outputs $\hat{y}$ from the agents, the belief state variable $\hat{\pi}_t^i=p(x|\hat{Y}^i)$ can be written as 
     \begin{equation}
\begin{split}
\hat{\pi}^1_t&=p(x_t|y^1_{t:t+h})  \qquad \qquad \qquad s^1_t=\gamma( \hat{\pi}^1_t) \\
\hat{\pi}^2_t&=p(x_t|y^1_{t:t+h}, y^2_{t:t+h}) \; \; \quad \qquad s^2_t=\gamma( \hat{\pi}^2_t) \\
\centerdot \;\;\, &= \quad \centerdot \qquad \centerdot   \qquad \qquad \qquad  \centerdot \qquad \centerdot  \\
\hat{\pi}^n_t &= p(x_t|y^1_{t:t+h},\cdots, y^n_{t:t+h}) \qquad  s^n_t=\gamma( \hat{\pi}^n_t) 
\end{split},
\end{equation} 
where $\hat{Y}^i$ denotes the set of expected outputs up to the $i$-th agent. In this way, the expected belief state can be updated sequentially based on the decisions of each agent, and can be used as known information for the $i+1$ agent. Here, $\gamma(*)$ is a decision making function given a target belief state whose output is desired path for each agent.
   
\subsection{Information-Theoretic Objective Path planning}
Our strategy is to maximize the information gain regarding the target belief state in a greedy fashion. Given each robot's expected information gains and travel costs, we obtain paths for all agents that maximize the overall utility.
\begin{equation}
\displaystyle \max_{\gamma^{1:n}} \mathbb{E}[\tilde{U}(x_t|y^1,y^2,\cdots,y^n)]
\end{equation}
where $\mathbb{E}[\tilde{U}$] refer to the expected value of the utility function given the target belief and the sensing outputs. This utility function can be calculated based on the path candidates for each agent in order to select the best path (or sequence of control inputs). Given a target belief at time $t$, obtaining the desired path for each agent allows MAS to gather information as quickly as possible. Our method aims to find the optimal path using a sampling-based optimization problem over a time horizon with an information-theoretic utility function. The utility function is described as 
\begin{equation}
%  \begin{split}
    \mathbb{E}[\tilde{U}(x_t|y^1,y^2,\cdots,y^n)] = \sum_{i}^{n}\left( IG(\text{s}_i)-c(\text{s}_i)\right)
    %   \end{split}
 \end{equation}
 
%   \begin{figure}[t]
% \label{Fig:python_environment}
% \includegraphics[width=1.0\columnwidth]{figures/vcd_trj.eps}
% \caption{Python simulation environment visualizing candidate local and global trajectories.}
% \end{figure}

\begin{algorithm}[t]
 \caption{Multi-Agent Search()}  \label{waypoint}
 \renewcommand{\algorithmicrequire}{\textbf{Input:}}
 \renewcommand{\algorithmicensure}{\textbf{Output:}}
  \begin{algorithmic}
  \Require {$Y$, $M$, $C$ \\ (Robot poses, Entropy Map, Sensing Capabilities (speed, coverage))}
  \Ensure {$S^*=\{s^*_1,s^*_2, \cdots, s^*_N\}$ (A set of paths)}
   \State $w_g \gets SampleWaypoints()$  \Comment Vertical Cell Decomposition
   \State $w_l \gets GetFrontiers()$  
%   \ForAll{$agents(n)$}
    \For{$n \gets 1$ to $N$} \Comment  for each agent
    \State $ s_g \gets SampleGlobalPaths(y_i, w_g)$ \Comment $A^*$ 
    \State $ s_l \gets SampleLocalPaths(y_i, w_l)$ \Comment $A^*$
    \State $S_i=\{s_g, s_l\}$ 
    \State $\hat{S_i}= reparameterize(S_i, C_i)$ \Comment  speed and coverage
    \For{$k \gets 1$ to $|S_i|$}                    
        \State {$IG(s_k) = CalculateIG(s_k,S^*_{1:n-1})$} \Comment Equation (7) \\ \Comment Use $S^*_{1:n-1}$ from higher hierarchy
        \State {$\bold{U}(s_k) = IG(s_k)-c(s_k)$} 
    \EndFor
    \State $s^{*} = s_k \gets \argmax{\bold{U}(s_k)}$ \Comment{Get the best path} 
    \State $S^*.append(s^*)$
      \EndFor \\ 
  \Return $S^{*}$ 
 \end{algorithmic}
 \end{algorithm}

  \begin{figure*}[t]
\centerline{\includegraphics[width=1.89\columnwidth]{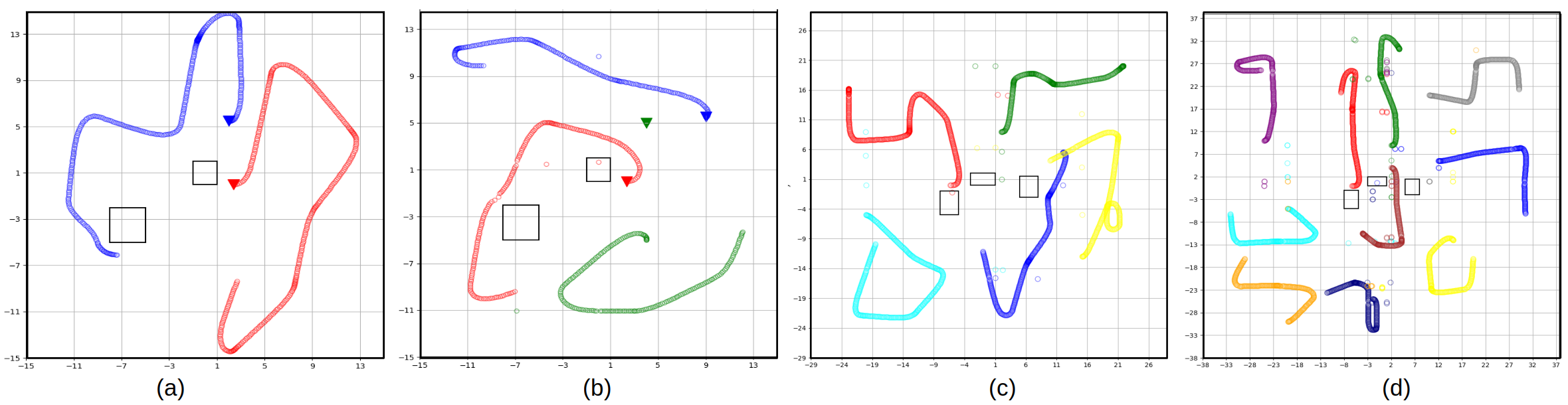}}
\caption{Simulation results using increasing number of agents. (a) 2 agents in a 13x13 grid. (b) 3 agents in a 13x13 grid. (c) 5 agents searching a 25x25 grid space. (d) 10 agents searching a 35x35 grid space.}
\label{fig:n_agent_search}
\end{figure*}

 \subsubsection{Path Selection}
To obtain the best path for each agent, we use a sampling-based optimization approach. The proposed strategy is to create global and local candidate paths. The goal points of paths can be sampled from waypoints within a search map boundary, sampled based on the entropy map $M$, which considers the target estimation model being updated by local measurements. Precisely, the global candidate points are sampled from a vertical cell decomposition of sub-regions whose occupancy grid type is unknown, $ M_{[m=0.5]}$. Cells with the value 0.5 are unknown while cells that are occupied or unoccupied are given 1.0 and 0.0, respectively. Local goal points are sampled from the cluster of frontiers \cite{yamauchi1997frontier}, which is defined as the boundary between known (occupied or free) and unknown areas, which are potentially informative. The frontiers can be obtained from the current entropy map. 
\begin{equation}
\begin{split}
p(x) &= \text{uniform over } M_{[m=0.5]} \\
\end{split}
\end{equation}
 Given the next viewpoint candidates and the current robot position, we use an A* planner to generate an obstacle-free path, $s$, for each agent. We note that any other planner like RRT or PRM could be used in its place. The set of paths to each global target point and a local target point are of different lengths and therefore must be re-parameterized. The reason for this re-parameterization is that the optimization problem should consider path length based on the robot's speed and the time horizon. In order to consider the difference in the speed of robots, the interval between the sampling points along the generated path is set accordingly. For example, if the path length $|\text{s}|$ is 10 and $ds=2$, we can sample five points along the trajectory. In order to avoid overlap between the computed FOV area, we sparsely sample points along a path to calculate the expected amount of information. Based on the number of sampled points along a path, we compute the Information Gain, denoted $IG(s)$, by the following equation:
 \begin{equation}
\begin{split}
     IG(s) \approx  \sum_{i=1}^{N_{\text{s}}}H(FOV(\text{s}^i)) \qquad \qquad \qquad \qquad \qquad \qquad \qquad \qquad\\
=\sum_{i=1}^{N_{\text{s}}}\left[\sum_{j=1}^{N_c}\left(p(m_{i,j})log(p(m_{i,j})) 
 +(1-p(m_{i,j}))log(1-p(m_{i,j})) \right) \right] 
\end{split}
\end{equation}
, where $\text{s}^i$ denotes the $i$-th sampled point and $p(m)$ is the occupancy probability, while $N_{\text{s}}$ and $N_c$ are the number of sampling points and the number of cells in the FOV given the sampled points, respectively. Thus, The $IG(\text{s})$ is calculated by summing over the FOV regions defined by sampled points through the trajectory. When calculating the information gain, the paths of the robots do not overlap as much as possible by not calculating the information gain corresponding to the path selected from the agent in the upper hierarchy. The proposed search algorithm is described in the Algorithm 1. 
 
\begin{figure}[t]
\centerline{\includegraphics[width=0.87\columnwidth]{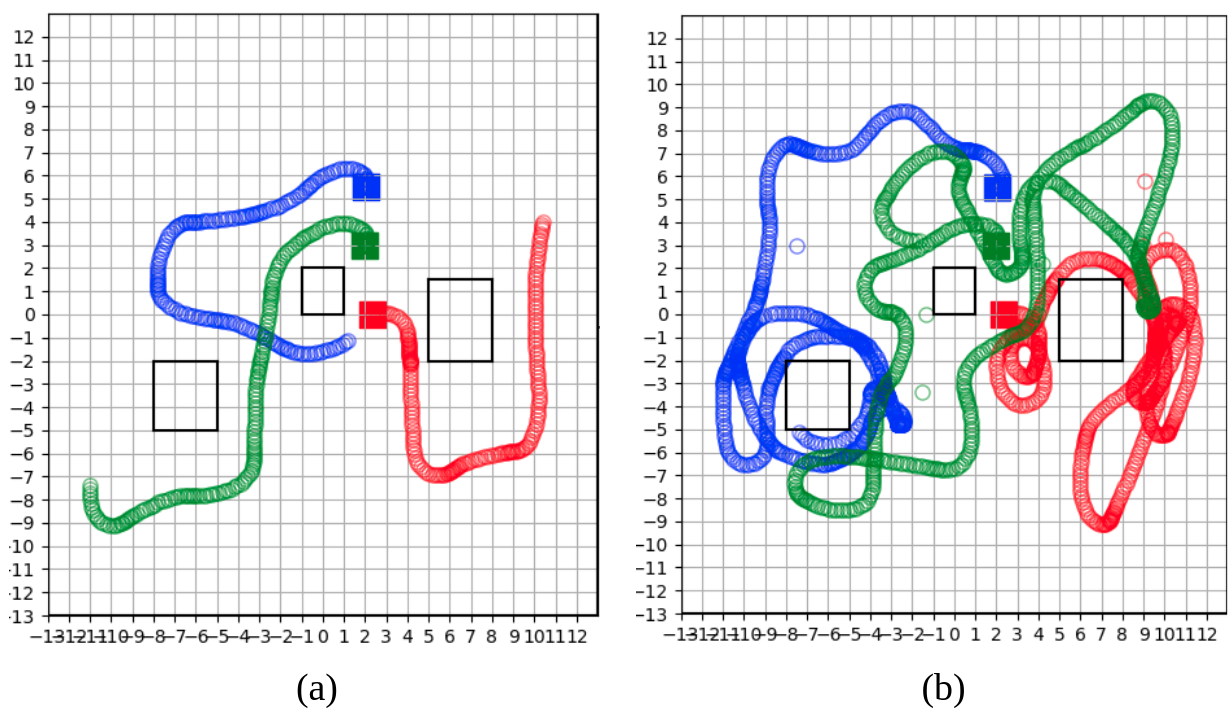}}
\caption{Simulation results with different conditions. (a) One time search (static target) (b) Continuous search (dynamic target).}
\label{fig:tworobot_search}
\end{figure}

 \begin{figure}
\centerline{\includegraphics[width=0.87\columnwidth]{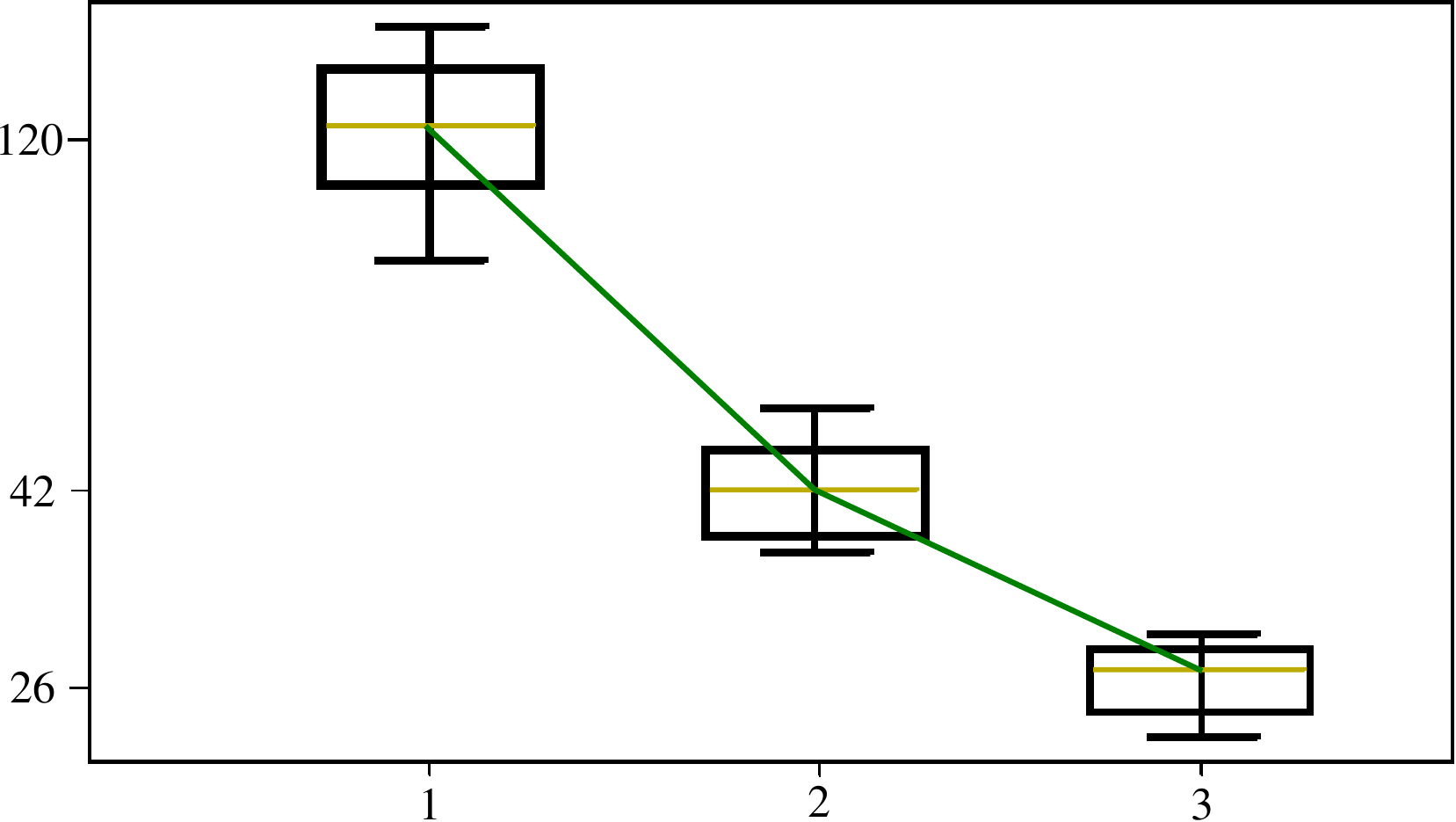}}
\caption{Average search time with varying number of agents.}
\label{fig:search_time}
\end{figure}

\begin{figure*}
\centerline{\includegraphics[width=1.75\columnwidth]{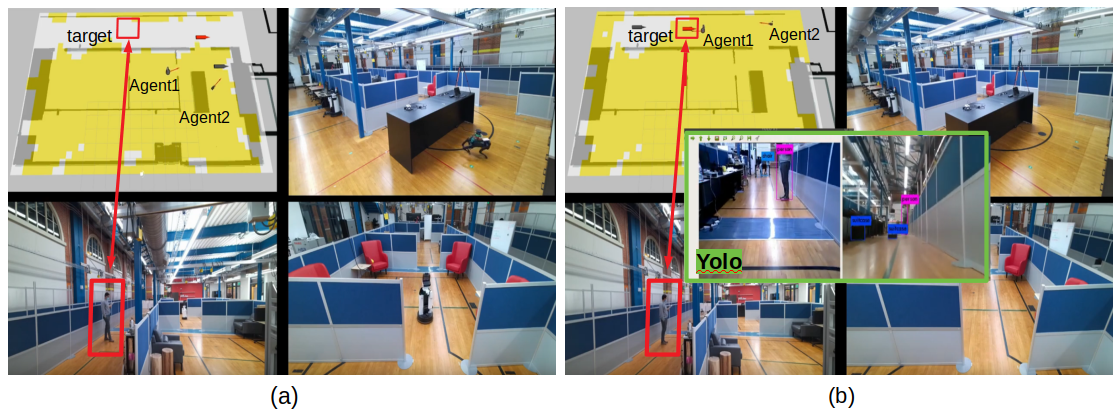}}
\caption{(a) Experimental validation for a two robot search in the Anna Hiss Gymnasium apartment area. In the top left figure, the yellow regions are those which have been explored, while the white areas are regions of uncertainty. The target location is described as a red box and is unknown to the robots. The red and black markers represent the next waypoints for each agent. (b) The completion of the experimental validation with both agents converging to the target of interest. The object recognition is performed using the well-known YOLO algorithm to detect people (shown in the green box).}
\label{fig:noYOLO}
\end{figure*}

\section{Results}
\subsection{Simulation Results}
In this section we present python-based simulation results of our proposed approach for a multi-agent search behavior and demonstrate the scalability of the algorithm. It is assumed that the simulation environment (search region) and all the static obstacles have a rectangular shape and obstacles are not known in advance. Each agent is equipped with a simulated ray sensor, which has a square field of view with limited range, $F(x,y)$, determined by the 2D position of the agent. Given the resolution of the map, the entire environment can be decomposed into square-type grid cells. To achieve robust collision avoidance, we use a dynamic window approach to generate the control input to navigate towards goal points.

\subsubsection{Single Search vs Continuous Search}
In order to test the search performance, simulations were conducted under various initial conditions. Depending on how the search map is updated (time-varying condition), we can implement one time search (similar to the exploration and mapping problem) and a continuous search. As shown in Fig \ref{fig:tworobot_search}, in the case of three agents, we can prove that the agents were able to effectively search for the target over the search region.% TODO: Can we prove it..? 
  
 \subsubsection{n-Agent Case}
 To validate the scalability of the proposed method, we test the exploration with $n$ agents. As the number of agents increases, we enlarge the search space. Fig. \ref{fig:n_agent_search} demonstrates the trajectories of each agent in the case number of agents equals to 2,3,5, and 10. Each color represents the trajectory of a different agent. These results demonstrate that our algorithm is scalable and can be extended to a general multi-agent system and still perform in real-time. 

 \begin{figure}
\centerline{\includegraphics[width=0.85 \columnwidth]{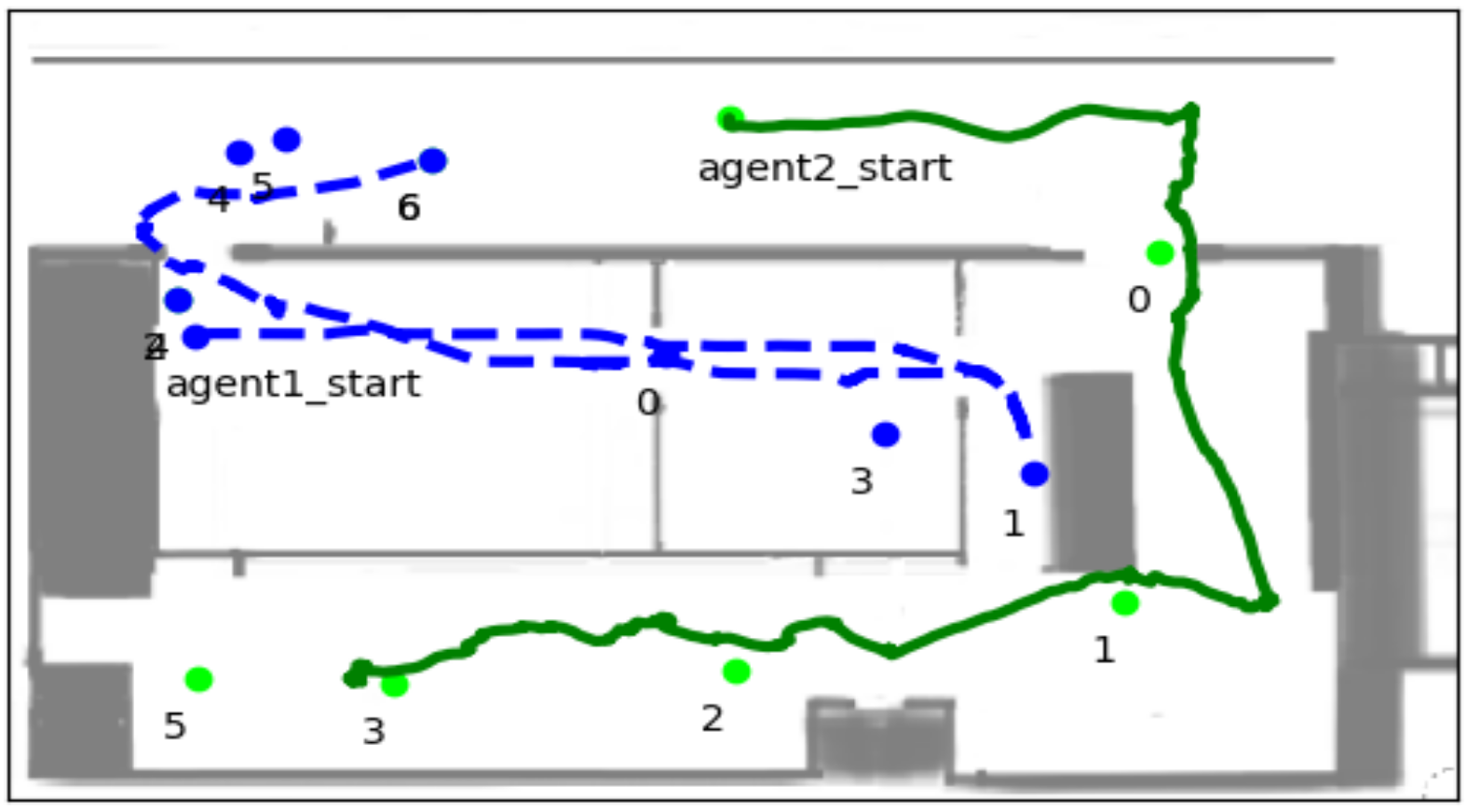}}
\caption{Trajectories for each agent in the AHG apartment setup with different sets of initial conditions}
\label{fig:ahg_trj}
\end{figure}

 \subsubsection{Search Time}
The search time varies depending on the initial condition or the dimension of the search space. Therefore, for accurate comparison, given a fixed size of the search map(13x13), fixed maximum moving speed (1m/s), the search time was compared. Fig. \ref{fig:search_time} shows the average search time (the entropy reduction rate) for three cases. By adding an additional agent, the time to completion is reduced by more than half.

\subsection{Experimental Results}

We used the Unitree A1 quadruped and the Toyota HSR mobile manipulation robot for experimentation of multiple agent target search. The A1 is equipped with a Velodyne VLP-16 3D Lidar and a RealSense D435 camera. On-board computing is performed in an Intel NUC Mini PC, which communicates with the low-level control systems. The HSR is equipped with a Hokuyo 2D Lidar, a RGB-D camera, and an on-board Jetson TK1 GPU. We assume that both robots have perfect localization, although in practice we provide it via Episodic non-Markov Localization \cite{biswas2017episodic}. The use of such different systems demonstrates that our algorithm is well suited to perform with a heterogeneous team, each with varying motion models. 

Experiments were performed in the Anna Hiss Gymnasium apartment at the University of Texas. Fig.~\ref{fig:noYOLO} shows two different moments in the search. The search map is 20(m) x 10(m) and the maximum velocity of each agent are 0.3(m/s) and 1.0(m/s) for the HSR and A1, respectively. In the top left of Fig.~\ref{fig:noYOLO} (a) and (b) is the global entropy map. Regions in yellow indicate that they have been explored (known to be free or occupied), while regions in white have high uncertainty. In the lower-left of each figure (a) and (b), the target is found in the hallway along the left side of the apartment setup. The red and black markers indicate the next waypoints for each agent, which is determined by using the search server. In (b), the object detection feed is shown as the two agents approach the target of interest. Finally, Fig.~\ref{fig:ahg_trj} shows resulting trajectories for the two agents for one of experimental trials. The video demo can be found at https://youtu.be/7WMqG7EiUVY.

\section{Conclusion}
\label{sec:conclusion}
This paper addresses online search for a heterogeneous multi-agent system. We employ an information-theoretic utility function and sampling-based optimization to obtain each agent's path. A hierarchical decision-making structure allows us to reduce computational burden and perform the search in real-time. Simulation results show that our proposed algorithm proves its scalability and that it can be extended to the general case of multiple agents. We further validate this algorithm by implementing it in a real-world environment. Overall results validate the effectiveness and robustness of the proposed method. 

% Based on the proposed framework, several future works remain. First, we will aim to search for a specific target within small crowds of people. The most representative example is an application that finds a specific person in a crowd within a search region using multiple robot agents. Adopting different target motion models such as moving targets or an adversarial target, we will verify that the proposed search behavior is effective and robust to that setting. Lastly, we plan to study whether a better solution can be found through a data-driven approach so that our agents can learn how to search by interacting with the environment. 

\section*{Acknowledgments}
The authors would like to thank the members of the Human Centered Robotics Laboratory at The University of Texas at Austin for their great help and support. This work was supported by the Army Futures Command and the Office of Naval Research, ONR Grant \#N000141512507.

%% Use plainnat to work nicely with natbib. 
\bibliographystyle{plainnat}
\bibliography{main}

\begin{thebibliography}{32}
\providecommand{\natexlab}[1]{#1}
\providecommand{\url}[1]{\texttt{#1}}
\expandafter\ifx\csname urlstyle\endcsname\relax
  \providecommand{\doi}[1]{doi: #1}\else
  \providecommand{\doi}{doi: \begingroup \urlstyle{rm}\Url}\fi

\bibitem[Agmon et~al.(2006)Agmon, Hazon, and Kaminka]{agmon2006constructing}
Noa Agmon, Noam Hazon, and Gal~A Kaminka.
\newblock Constructing spanning trees for efficient multi-robot coverage.
\newblock In \emph{Proceedings 2006 IEEE International Conference on Robotics
  and Automation, 2006. ICRA 2006.}, pages 1698--1703. IEEE, 2006.

\bibitem[Amigoni and Caglioti(2010)]{caglioti2010}
Francesco Amigoni and Vincenzo Caglioti.
\newblock An information-based exploration strategy for environment mapping
  with mobile robots.
\newblock \emph{Robotics and Autonomous Systems}, 58\penalty0 (5):\penalty0
  684--699, 2010.

\bibitem[Aydemir et~al.(2011)Aydemir, Sj{\"o}{\"o}, Folkesson, Pronobis, and
  Jensfelt]{aydemir2011search}
Alper Aydemir, Kristoffer Sj{\"o}{\"o}, John Folkesson, Andrzej Pronobis, and
  Patric Jensfelt.
\newblock Search in the real world: Active visual object search based on
  spatial relations.
\newblock In \emph{2011 IEEE International Conference on Robotics and
  Automation}, pages 2818--2824. IEEE, 2011.

\bibitem[Aydemir et~al.(2013)Aydemir, Pronobis, G{\"o}belbecker, and
  Jensfelt]{aydemir2013active}
Alper Aydemir, Andrzej Pronobis, Moritz G{\"o}belbecker, and Patric Jensfelt.
\newblock Active visual object search in unknown environments using uncertain
  semantics.
\newblock \emph{IEEE Transactions on Robotics}, 29\penalty0 (4):\penalty0
  986--1002, 2013.

\bibitem[Bellicoso et~al.(2018)Bellicoso, Bjelonic, Wellhausen, Holtmann,
  G{\"u}nther, Tranzatto, Fankhauser, and Hutter]{bellicoso2018advances}
C~Dario Bellicoso, Marko Bjelonic, Lorenz Wellhausen, Kai Holtmann, Fabian
  G{\"u}nther, Marco Tranzatto, Peter Fankhauser, and Marco Hutter.
\newblock Advances in real-world applications for legged robots.
\newblock \emph{Journal of Field Robotics}, 35\penalty0 (8):\penalty0
  1311--1326, 2018.

\bibitem[Biswas and Veloso(2017)]{biswas2017episodic}
Joydeep Biswas and Manuela~M Veloso.
\newblock Episodic non-markov localization.
\newblock \emph{Robotics and Autonomous Systems}, 87:\penalty0 162--176, 2017.

\bibitem[Charrow et~al.(2015{\natexlab{a}})Charrow, Kahn, Patil, Liu, Goldberg,
  Abbeel, Michael, and Kumar]{kumarplan2015}
Benjamin Charrow, Gregory Kahn, Sachin Patil, Sikang Liu, Ken Goldberg, Pieter
  Abbeel, Nathan Michael, and Vijay Kumar.
\newblock Information-theoretic planning with trajectory optimization for dense
  3d mapping.
\newblock In \emph{Robotics: Science and Systems}, volume~11,
  2015{\natexlab{a}}.

\bibitem[Charrow et~al.(2015{\natexlab{b}})Charrow, Liu, Kumar, and
  Michael]{kumarmap2015}
Benjamin Charrow, Sikang Liu, Vijay Kumar, and Nathan Michael.
\newblock Information-theoretic mapping using cauchy-schwarz quadratic mutual
  information.
\newblock In \emph{2015 IEEE International Conference on Robotics and
  Automation (ICRA)}, pages 4791--4798. IEEE, 2015{\natexlab{b}}.

\bibitem[Dai et~al.(2020)Dai, Papatheodorou, Funk, Tzoumanikas, and
  Leutenegger]{dai2020}
Anna Dai, Sotiris Papatheodorou, Nils Funk, Dimos Tzoumanikas, and Stefan
  Leutenegger.
\newblock Fast frontier-based information-driven autonomous exploration with an
  mav.
\newblock \emph{arXiv preprint arXiv:2002.04440}, 2020.

\bibitem[Dang et~al.(2020)Dang, Tranzatto, Khattak, Mascarich, Alexis, and
  Hutter]{dang2020graph}
Tung Dang, Marco Tranzatto, Shehryar Khattak, Frank Mascarich, Kostas Alexis,
  and Marco Hutter.
\newblock Graph-based subterranean exploration path planning using aerial and
  legged robots.
\newblock \emph{Journal of Field Robotics}, 37\penalty0 (8):\penalty0
  1363--1388, 2020.

\bibitem[Erik J.~Forsmo et~al.(2013)Erik J.~Forsmo, Thor I.~Fossen, and
  Arne]{milp2013}
Esten I.~Grotli Erik J.~Forsmo, Johansen Thor I.~Fossen, and Tor Arne.
\newblock Optimal search mission with unmanned aerial vehicles using mixed
  integer linear programming.
\newblock 2013.

\bibitem[Filatov and Krinkin(2020)]{filatov2020simplistic}
Anton Filatov and Kirill Krinkin.
\newblock A simplistic approach for lightweight multi-agent slam algorithm.
\newblock \emph{International Journal of Embedded and Real-Time Communication
  Systems (IJERTCS)}, 11\penalty0 (3):\penalty0 67--83, 2020.

\bibitem[G{\"o}belbecker et~al.(2011)G{\"o}belbecker, Aydemir, Pronobis,
  Sj{\"o}{\"o}, and Jensfelt]{gobelbecker2011planning}
Moritz G{\"o}belbecker, Alper Aydemir, Andrzej Pronobis, Kristoffer
  Sj{\"o}{\"o}, and Patric Jensfelt.
\newblock A planning approach to active visual search in large environments.
\newblock In \emph{Automated Action Planning for Autonomous Mobile Robots},
  2011.

\bibitem[Gonzalez et~al.(2005)Gonzalez, Alvarez, Diaz, Parra, and
  Bustacara]{gonzalez2005bsa}
Enrique Gonzalez, Oscar Alvarez, Yul Diaz, Carlos Parra, and Cesar Bustacara.
\newblock Bsa: a complete coverage algorithm.
\newblock In \emph{Proceedings of the 2005 IEEE International Conference on
  Robotics and Automation}, pages 2040--2044. IEEE, 2005.

\bibitem[Grotli and Johansen(2012)]{milp2012}
Esten~Ingar Grotli and Tor~Arne Johansen.
\newblock Path planning for uavs under communication constraints using splat!
  and milp.
\newblock pages 265--282, 2012.

\bibitem[Kan et~al.(2020)Kan, Teng, and Kayrdis]{karydis2020}
Xinyue Kan, Hanzhe Teng, and Konstantinos Kayrdis.
\newblock Online exploration and coverage planning in unknown
  obstacle-cluttered environments.
\newblock 2020.

\bibitem[Khan et~al.(2017)Khan, Noreen, Ryu, Doh, and Habib]{proximitycpp2017}
Amna Khan, Iram Noreen, Hyejeong Ryu, Nakju~Lett Doh, and Zulfiqar Habib.
\newblock Online complete coverage path planning using two-way proximity
  search.
\newblock pages 229--240, 2017.

\bibitem[Kontitsis et~al.(2013)Kontitsis, Theodorou, and Todorov]{todorov2013}
Michail Kontitsis, Evangelos~A Theodorou, and Emanuel Todorov.
\newblock Multi-robot active slam with relative entropy optimization.
\newblock In \emph{2013 American Control Conference}, pages 2757--2764. IEEE,
  2013.

\bibitem[Krinkin et~al.(2017)Krinkin, Filatov, and Filatov]{slamsurvey2017}
Kirill Krinkin, Anton Filatov, and Artyom Filatov.
\newblock Modern multi-agent slam approaches survey.
\newblock In \emph{Proceedings of the XXth Conference of Open Innovations
  Association FRUCT}, volume 776, pages 617--623, 2017.

\bibitem[Lee et~al.(2020)Lee, Kim, and Lee]{lee2020efficient}
SeungHwan Lee, HanJun Kim, and BeomHee Lee.
\newblock An efficient rescue system with online multi-agent slam framework.
\newblock \emph{Sensors}, 20\penalty0 (1):\penalty0 235, 2020.

\bibitem[Rasouli and Tsotsos(2016)]{rasouli2016sensor}
Amir Rasouli and John~K Tsotsos.
\newblock Sensor planning for 3d visual search with task constraints.
\newblock In \emph{2016 13th Conference on Computer and Robot Vision (CRV)},
  pages 37--44. IEEE, 2016.

\bibitem[Richard~Bormann and Hagele(2018)]{indoorsurvey2018}
Joshua~Hampp Richard~Bormann, Florian~Jordan and Martin Hagele.
\newblock Indoor coverage path planning: Survey, implementation, analysis.
\newblock 2018.

\bibitem[Scherer et~al.(2015)Scherer, Yahyanejad, Hayat, Yanmaz, Andre, Khan,
  Vukadinovic, Bettstetter, Hellwagner, and Rinner]{scherer2015autonomous}
J{\"u}rgen Scherer, Saeed Yahyanejad, Samira Hayat, Evsen Yanmaz, Torsten
  Andre, Asif Khan, Vladimir Vukadinovic, Christian Bettstetter, Hermann
  Hellwagner, and Bernhard Rinner.
\newblock An autonomous multi-uav system for search and rescue.
\newblock In \emph{Proceedings of the First Workshop on Micro Aerial Vehicle
  Networks, Systems, and Applications for Civilian Use}, pages 33--38, 2015.

\bibitem[Schmuck and Chli(2017)]{schmuck2017multi}
Patrik Schmuck and Margarita Chli.
\newblock Multi-uav collaborative monocular slam.
\newblock In \emph{2017 IEEE International Conference on Robotics and
  Automation (ICRA)}, pages 3863--3870. IEEE, 2017.

\bibitem[Shubina and Tsotsos(2010)]{shubina2010visual}
Ksenia Shubina and John~K Tsotsos.
\newblock Visual search for an object in a 3d environment using a mobile robot.
\newblock \emph{Computer Vision and Image Understanding}, 114\penalty0
  (5):\penalty0 535--547, 2010.

\bibitem[Song and Gupta(2018)]{songcpp2018}
Junnan Song and Shalabh Gupta.
\newblock Epsilon*: An online coverage path planning algorithm.
\newblock pages 526--533, 2018.

\bibitem[Song and Gupta(2019)]{song2019}
Junnan Song and Shalabh Gupta.
\newblock Care: Cooperative autonomy for resilience and efficiency of robot
  teams for complete coverage of unknown environments under robot failure.
\newblock 2019.

\bibitem[Wang et~al.(2020)Wang, Zhang, Liang, Chen, and Wang]{wang2020multi}
Tian-miao Wang, Yi-cheng Zhang, Jian-hong Liang, Yang Chen, and Chao-lei Wang.
\newblock Multi-uav collaborative system with a feature fast matching
  algorithm.
\newblock \emph{Frontiers of Information Technology \& Electronic Engineering},
  pages 1--18, 2020.

\bibitem[Woosley et~al.(2020)Woosley, Dasgupta, Rogers, and
  Twigg]{woosley2020multi}
Bradley Woosley, Prithviraj Dasgupta, John~G Rogers, and Jeffrey Twigg.
\newblock Multi-robot information driven path planning under communication
  constraints.
\newblock \emph{Autonomous Robots}, 44\penalty0 (5):\penalty0 721--737, 2020.

\bibitem[Yamauchi(1997)]{yamauchi1997frontier}
Brian Yamauchi.
\newblock A frontier-based approach for autonomous exploration.
\newblock In \emph{Computational Intelligence in Robotics and Automation, 1997.
  CIRA'97., Proceedings., 1997 IEEE International Symposium on}, pages
  146--151. IEEE, 1997.

\bibitem[Ye and Tsotsos(1999)]{ye1999sensor}
Yiming Ye and John~K Tsotsos.
\newblock Sensor planning for 3d object search.
\newblock \emph{Computer Vision and Image Understanding}, 73\penalty0
  (2):\penalty0 145--168, 1999.

\bibitem[Zhang and Sridharan(2012)]{zhang2012active}
Shiqi Zhang and Mohan Sridharan.
\newblock {Active visual sensing and collaboration on mobile robots using
  hierarchical POMDPs}.
\newblock In \emph{Proceedings of the 11th International Conference on
  Autonomous Agents and Multiagent Systems-Volume 1}, pages 181--188.
  International Foundation for Autonomous Agents and Multiagent Systems, 2012.

\end{thebibliography}
% \bibliographystyle{IEEEtran}
% \bibliography{references}

\end{document}